\documentclass[11pt]{article}

\usepackage[margin=1in]{geometry}
\usepackage{amsmath}
\usepackage{amssymb}
\usepackage{booktabs}
\usepackage{graphicx}
\usepackage{xcolor}
\usepackage{caption}
\usepackage{enumitem}
\usepackage[colorlinks=true,linkcolor=blue,urlcolor=blue,citecolor=blue]{hyperref}

\newcommand{\code}[1]{\texttt{#1}}
\title{Does a Language Server Save Tokens for Coding Agents?\\
\large A Measurement Methodology and Preliminary Study}
\author{Pengcheng Xu \\ \small \texttt{pooytr1@gmail.com}}
\date{June 2026}

\begin{document}
\maketitle

\begin{abstract}
Coding agents spend most of their context budget on \emph{retrieval}: deciding which tokens of a
repository must enter the model's context so it can build a faithful internal model of the task. Two
regimes dominate. \textbf{Lexical retrieval} (\code{grep}, \code{ripgrep}, \code{find}) is universal,
instant, and zero-setup, but noisy: it cannot distinguish a definition from a call from a mention in a
comment. \textbf{Semantic retrieval} via the Language Server Protocol (LSP) ---
\code{textDocument/references}, \code{definition}, \code{hover}, \code{documentSymbol} --- is precise
and typed, but requires a running, indexed server and pays a per-symbol round-trip cost. The widely
repeated claim is that semantic retrieval is ``more token-efficient.'' We surveyed the tools and
literature that make this claim and found a striking gap: \textbf{it is asserted almost everywhere and
measured almost nowhere.} This paper (1) formalizes the question with a single primary metric,
\emph{tokens-to-success}; (2) specifies a five-arm ablation isolating semantic retrieval from
confounds; (3) maps three pre-stated failure modes onto measurable variables; and (4) reports a
\textbf{preliminary measurement study} across four task classes (Python \code{requests} plus two
TypeScript repositories; Claude Opus~4.8, Sonnet~4.6, Haiku~4.5) that turns the prior into first
numbers. The answer is conditional and, in the common case,
\textbf{negative}: on symbol-named \emph{localization} the LSP \emph{costs} tokens ($+6\%$ Opus,
$+118\%$ Sonnet) and the agent ignores it entirely when free; on \emph{reference-completeness} the LSP
buys \textbf{precision} ($1.00$ vs $0.76$, zero false call sites) but not token savings (a ${\sim}19\%$
premium) and cannot raise the recall ceiling set by agent thoroughness; and it nets a token
\textbf{saving only for the weakest model} (Haiku, $-26\%$), as a crutch against lexical noise. On
\emph{editing} scored by real test execution the verdict is sharpest: \code{grep} solves multi-file
renames perfectly while a location-only LSP fails three-quarters of them by missing a call site, and even
a complete, index-warmed, text-enriched LSP (returning each reference's line inline, as production
LSP-MCP servers do) recovers most of that gap --- cutting follow-up file-reads ${\sim}5\times$ --- but
cannot close it, because a rename must touch comments and strings that semantic references structurally
exclude. The most robust finding is that the agent's tool choice is \textbf{task-dependent}: it defaults to
\code{grep} on localization (semantic-tool use $0$--$6\%$) but reaches for the LSP about half the time on
reference tasks ($45$--$57\%$), \emph{unprompted}. The grep preference is not a fixed bias but a learned,
task-shaped policy. This argues not for ``LSP-always'' but for an \emph{adaptive router} keyed on task
class, model capability, and lexical noise --- and, since that routing competence is demonstrably already
present in latent form, for reinforcing it into the policy rather than bolting on the tool.
\end{abstract}

\section{The fundamental question}

An agent navigating a codebase is, in essence, performing \textbf{retrieval under a token budget}. The
model has a finite context window; the task is solvable only if the \emph{right} tokens enter that
window and the \emph{wrong} tokens stay out. Every tool call is a retrieval decision with a price,
denominated in tokens.

From this lens, the two regimes are two points on a precision/recall/cost surface. \textbf{Lexical
retrieval} has high recall and low precision: it finds every textual match, including matches in
comments, strings, and unrelated identifiers, and the agent often pays again to read surrounding lines.
\textbf{Semantic retrieval} has high precision: ``find references'' returns exactly the resolved
references. The cost moves elsewhere --- a server must run and index, and each query is a JSON-RPC
round-trip whose latency grows with repository size. So the question has a precise form:

\begin{quote}
\textit{At equal task-success rate, how many fewer tokens does semantic retrieval place into the
agent's context than lexical retrieval, and under what conditions does that delta become negative?}
\end{quote}

The phrase \emph{at equal task-success rate} is load-bearing: a method that ``saves tokens'' by failing
earlier has saved nothing. Token savings are meaningful only \textbf{conditioned on iso-accuracy}.

\section{Background and related work}

\paragraph{Provenance note.} Live web \emph{search} tooling was degraded during this study (returning
model-generated synthesis rather than ranked results). Two retrieval paths returned verifiable primary
sources, and every claim below is grounded in one: \textbf{(F)} fetch-verified this session (the OSS
tools below, and arXiv papers retrieved via the arXiv API with full abstracts); \textbf{(K)} established
background from training knowledge, flagged where used and never relied on for a novel quantitative
claim. This separation is deliberate: the paper's central claim is about the \emph{absence} of
measurement, so it must not itself rest on unverifiable measurement.

\paragraph{Tools that already expose semantic navigation (F).}
\textbf{Serena}~\cite{serena} wraps LSP with \code{find\_symbol},
\code{find\_referencing\_symbols}, and symbolic editing across 40+ languages; it states symbolic editing
is ``much more token-efficient'' and lets an agent explore ``without reading entire files'' --- an
explicit claim \emph{with no benchmark numbers}. \textbf{mcp-language-server}~\cite{mcpls} wraps a real language server and calls its edit
tool ``more reliable and context-economical'' --- scoped to editing, \emph{no general token comparison}.
\textbf{Aider repo-map}~\cite{aidermap} is a static cousin of LSP:
tree-sitter extracts signatures and a PageRank-style ranking keeps the most-referenced symbols within a
token budget; its stated motivation is exactly our thesis (``sending whole files \dots\ waste[s] the
precious context window'') but it reports \emph{no before/after numbers}.

\paragraph{Recent literature (F, arXiv, full abstracts captured).}
\textbf{TypeScript Repository Indexing for Code Agent Retrieval}~\cite{tsindex} --- most directly
relevant, in our originally-chosen language. It observes that ``LSP-based resolution requires a JSON-RPC
call for each symbol lookup, [so] these per-symbol calls become a bottleneck on large TypeScript
repositories,'' motivating a TS-Compiler-API parser. It measures \emph{index-construction} efficiency,
not an agent's tokens-to-success. \textbf{RL from Compiler and Language Server Feedback}~\cite{rlcsf} argues compilers/type-checkers/language-servers ``already compute the missing
supervision signal \dots\ but expose it through interfaces designed for human-driven IDEs rather than
learning loops,'' and turns LSP signal into a shaped process reward --- the strongest articulation of
the interface-friction hypothesis and of training the signal into the policy. \textbf{CORE-Bench}~\cite{corebench} is a benchmark for requirement-driven repository search (180K+ queries on
SWE-bench-series), reporting ``a sharp drop from traditional code search to code retrieval in agentic
coding settings.'' \textbf{Code as Agent Harness}~\cite{codeharness} frames code/tooling as the agent's
substrate and lists ``evaluation beyond final task success'' as an open challenge.

\paragraph{The gap.} Across tools and papers the pattern is consistent: \textbf{the token-efficiency of
semantic retrieval is asserted, not measured.} The closest quantitative work measures index-build
efficiency, not agent tokens-to-success at iso-accuracy. That gap is this study's justification.
Established background (K): SWE-bench~\cite{swebench} and the SWE-agent agent-computer-interface
line~\cite{sweagent} provide the verified-task harness our protocol reuses.

\section{Method}

\paragraph{Primary metric: tokens-to-success.} Let a \emph{task} be a repository state plus a
verifiable goal. For agent configuration $c$ on task $t$, run $R$ rollouts. Define
\[
\mathrm{success}(c,t) = \frac{1}{R}\sum_{r=1}^{R} \mathbb{1}[\text{rollout } r \text{ verified}],
\qquad
\mathrm{T2S}(c,t) = \frac{\sum_{r:\,\text{verified}} \mathrm{tokens}(c,t,r)}{\#\{r : \text{verified}\}},
\]
where $\mathrm{tokens}(c,t,r)$ is the \textbf{total context tokens} consumed by rollout $r$ (prompt +
tool-result + generated, summed over turns --- what the user pays for). The headline comparison is
always the pair $(\text{success rate}, \text{T2S})$; we never report a token number without its success
rate. Secondary, per-rollout: tool-call counts (by tool), read volume, turns, wall-clock, and the
\textbf{grep false-positive rate} (fraction of lexical matches that are not semantically relevant ---
the theoretical headroom for precision).

\paragraph{The ablation: five arms.} All arms share the \textbf{same model, tasks, harness, and prompt
scaffold}; the only variable is the retrieval tool surface.

\begin{center}\small
\begin{tabular}{@{}lll@{}}
\toprule
Arm & Tool surface & Purpose \\
\midrule
A grep-only & \code{grep}+read+edit & status quo \\
B lsp-only & \code{references}/\code{definition}/\code{documentSymbol}+read+edit & semantic in isolation \\
C both & A $\cup$ B (agent chooses) & realistic; reveals revealed preference \\
D forced-semantic & A $\cup$ B, prompt mandates semantic-first & habit vs.\ capability \\
E static repo-map & Aider-style signature map + grep & needs a \emph{live} server? \\
\bottomrule
\end{tabular}
\end{center}

The A--B--C triangle answers ``does semantic help when available, and does the agent use it when free?''
D vs.\ C isolates \emph{habit} from \emph{capability}; E vs.\ B isolates how much benefit needs a live
server. (Arm E is untested here.)

\paragraph{Tasks and controls.} We use tasks where reference-finding is on the critical path:
change-signature/fix-callers, who-calls-X, cross-file refactor, dead-code, and issue-to-edit
localization, drawn from SWE-bench-series so success is objectively checkable. We hold model and decoding
fixed, match tool-description length across arms, measure LSP arms both warm and cold, and log every
tool call with its token cost so deltas can be attributed to (a) fewer reads, (b) less noise per read,
or (c) fewer turns.

\section{The three failure modes, made measurable}

@pchsu pre-stated three reasons an LSP might fail in practice. Each maps to a measurement and a remedy.
\textbf{H-install} (hard to install/initialize) $\rightarrow$ measure per-language setup success and
cold-start cost; remedy arm: \emph{index-on-build} (for TS, via the Compiler API rather than a live
session, following arXiv:2604.18413). \textbf{H-interface} (API too limited) $\rightarrow$ log every
\emph{LSP-insufficiency event} where the agent falls back to grep, and categorize; the taxonomy of
fallbacks is itself an output pointing at candidate API extensions. \textbf{H-churn} (edits invalidate
the index) $\rightarrow$ measure re-index latency and ``tokens wasted re-reading after edit''; remedy
arm: partial-index + grep hybrid. A clean negative on any remedy is a real finding --- it tells a team
where \emph{not} to invest.

\section{Why agents prefer \code{grep} --- mechanism hypotheses}

Given a free choice, agents reach for \code{grep} and rarely issue LSP queries. Ranked by prior:
(1)~\textbf{Training prior} --- pretraining/RLHF corpora are saturated with humans running \code{grep} in
terminals and nearly devoid of programmatic LSP JSON-RPC; the policy has a high prior on the
high-frequency tool. (2)~\textbf{Tool affordance} --- \code{grep} is universal and terse; LSP tools
carry higher perceived activation energy. (3)~\textbf{Universality} --- \code{grep} works in any repo
with zero setup. (4)~\textbf{Latency} --- \code{grep} returns instantly; a cold index stalls.
(5)~\textbf{Interface gaps} --- no single LSP call answers ``where is this \emph{conceptually} used''
across dynamic dispatch and strings. The decisive experiment is C vs.\ D: if forcing semantic-first
improves T2S over free choice, the agent was \emph{under-using} a capability it had (a habit problem);
if not, semantic retrieval genuinely was not better for those tasks (a coverage problem).

\section{Results}\label{sec:results}

We implemented the harness and ran the study on \textbf{Python / \code{requests}} (and two TypeScript
repositories for the cross-language and edit experiments) with two retrieval
servers (\code{python-lsp-server}, and \code{pyright} for the reference and edit experiments), driving Claude
models through a tool-use agent loop that logs token usage per turn. The study is preliminary --- small
task suites, few repositories --- and validates the methodology and surfaces first signals rather than settling the question at scale
(\S\ref{sec:limits}). All raw results are public (see Code and Data Availability).

\subsection{Localization --- grep wins on tokens}

Six SWE-bench-Lite \code{requests} issues; task = name the file(s) to change, verified against the gold
patch; four arms $\times$ 3 rollouts, Opus~4.8.

\begin{center}\small
\begin{tabular}{@{}lrrr@{}}
\toprule
Arm & success & T2S & free-choice semantic use \\
\midrule
A grep-only & 100\% & 920 & 53 grep / 0 semantic \\
B lsp-only & 100\% & \textbf{971 ($+6\%$)} & --- \\
C grep+lsp (free) & 100\% & 919 & 50 grep / \textbf{0 semantic} \\
D forced-semantic & 89\% & 945 & 51 grep / 4 semantic \\
\bottomrule
\end{tabular}
\end{center}

\begin{figure}[h]
\centering
\includegraphics[width=0.62\textwidth]{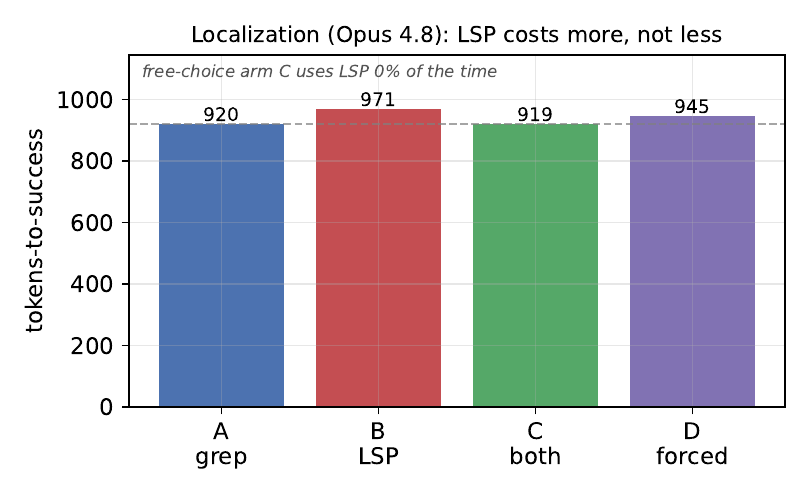}
\caption{\textbf{Localization (Opus 4.8): the LSP costs more, not less.} Tokens-to-success by arm. The
LSP-only arm (B) is $6\%$ more expensive than grep-only (A) at identical $100\%$ success. The
free-choice arm (C) collapses onto grep (it uses the LSP $0\%$ of the time), and forcing semantic-first
(D) \emph{reduces} success to $89\%$. On localization where the issue text names the symbol, semantic
retrieval is a token tax.}
\end{figure}

On localization where the issue text typically \emph{names} the symbol, \textbf{semantic retrieval costs
more tokens, not fewer}. Decisively, in arm C the agent used the LSP \textbf{zero times} --- C collapses
onto A. Forcing semantic-first (D) \emph{reduced} success and the agent still resisted (4 semantic vs.\
51 grep calls). This confirms \textbf{Prediction 2}, \emph{refutes} the localization half of
\textbf{Prediction 1} (for symbol-named localization, semantic retrieval is a tax), and confirms
\textbf{Prediction 3} emphatically: the habit gap is so strong that forcing backfires.

\subsection{Model sweep --- LSP helps weak models, taxes strong ones}

Same localization tasks, three models, 72 episodes each.

\begin{center}\small
\begin{tabular}{@{}lrrcr@{}}
\toprule
Model & grep T2S & lsp T2S & lsp vs.\ grep & free-choice semantic use \\
\midrule
Opus 4.8 & 920 & 971 & \textbf{$+6\%$} & 0\% \\
Sonnet 4.6 & 606 & 1{,}319 & \textbf{$+118\%$} & 4\% \\
Haiku 4.5 & 11{,}911 & 8{,}799 & \textbf{$-26\%$} & 6\% \\
\bottomrule
\end{tabular}
\end{center}

\begin{figure}[h]
\centering
\includegraphics[width=0.92\textwidth]{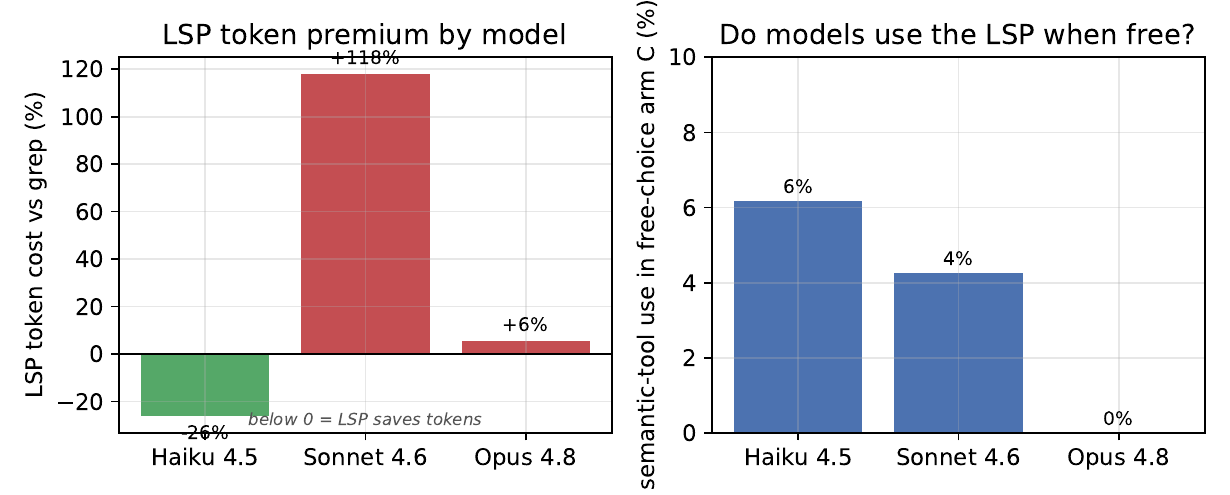}
\caption{\textbf{Capability dependence and the universal grep habit.} \emph{Left:} the LSP token premium
over grep, by model. Only the weakest model (Haiku) \emph{saves} tokens with the LSP ($-26\%$); both
capable models pay a premium (Opus $+6\%$, Sonnet $+118\%$). \emph{Right:} how often each model uses a
semantic tool in the free-choice arm C. Every model defaults to grep --- even Haiku, which would save
$26\%$, uses the LSP only $6\%$ of the time. Giving the tool is not enough.}
\end{figure}

The \textbf{capability dependence} is the headline: the weakest model (Haiku --- which flails with grep
noise, ${\sim}12$k tokens and 13+ turns per task) is the \emph{only} one that \textbf{saves} tokens with
the LSP; both capable models pay a premium. Semantic retrieval is a crutch for weak models and a tax for
strong ones. And the grep preference is \textbf{universal}: free-choice semantic use is $0\%/4\%/6\%$.

\subsection{Reference-completeness --- LSP buys precision, not tokens, and cannot fix recall}

Task = list \emph{every} call site of a target function, scored by F1 against \code{pyright}'s reference
set. (We switched the oracle from \code{pylsp} to \code{pyright} after finding jedi's reference
resolution too incomplete to serve as ground truth --- itself a finding about LSP quality variance.)
Five cross-file \code{requests} functions $\times$ 4 arms $\times$ 3 rollouts, Opus~4.8.

\begin{center}\small
\begin{tabular}{@{}lrrrr@{}}
\toprule
Arm & mean F1 & precision & recall & tokens \\
\midrule
A grep-only & 0.706 & 0.76 & 0.67 & 1136 \\
B lsp-only & \textbf{0.778} & \textbf{1.00} & 0.66 & 1347 ($+19\%$) \\
C grep+lsp (free) & 0.706 & 0.76 & 0.67 & 1354 \\
D forced-semantic & 0.710 & 0.77 & 0.67 & 1499 \\
\bottomrule
\end{tabular}
\end{center}

\begin{figure}[h]
\centering
\includegraphics[width=0.72\textwidth]{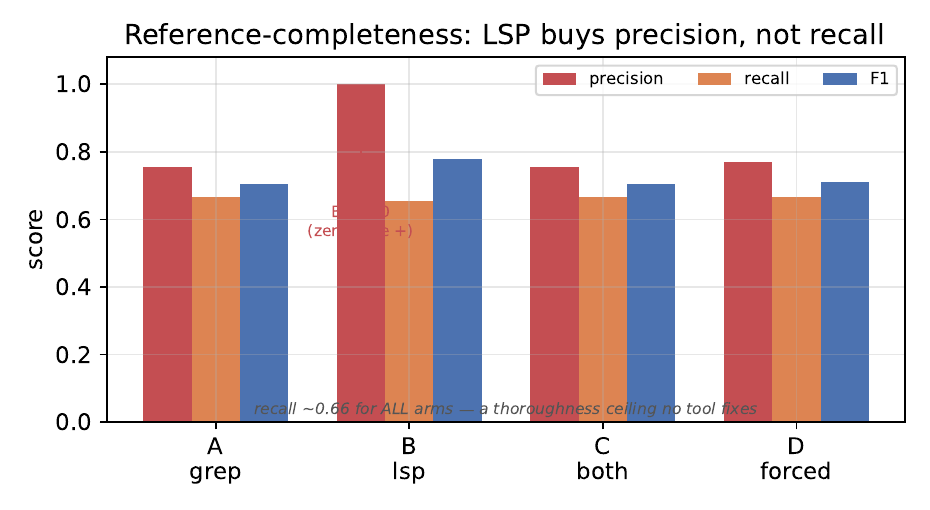}
\caption{\textbf{Reference-completeness: the LSP buys precision, not recall.} Precision, recall, and F1
by arm. The LSP arm (B) achieves \emph{perfect precision} (1.00 --- zero false call sites) versus grep's
0.76, and wins on F1. But \textbf{recall is identical ($\sim$0.66) across all arms}: no tool helps the
agent find \emph{more} true call sites. The missing third is an agent-thoroughness ceiling, not a
retrieval problem.}
\end{figure}

Here --- on \code{find\_references}' home turf --- \textbf{the LSP wins on accuracy} (F1 $0.778$ vs.\
$0.706$), the \emph{opposite} of localization: the task class decides whether semantic retrieval helps.
The decomposition is the real result. The entire gain is \textbf{precision}: B reports zero false call
sites (1.00) vs.\ grep's 0.76 ($\sim$24\% of grep hits are false positives --- comments, strings,
unrelated same-named symbols). \textbf{Recall is identical ($\sim$0.66) across all four arms}: neither
tool helps the agent find \emph{more} true sites --- the missing third is an \emph{agent-thoroughness}
problem, not a retrieval one, which the LSP cannot fix. The precision gain costs $\sim$19\% more tokens,
appears on every cross-file target ($+0.06$ to $+0.13$ F1), and vanishes on the one single-file target
(both arms F1 $=1.00$). Free-choice arm C reverts to grep's profile --- the agent forgoes the precision
gain even where it exists.

\begin{figure}[h]
\centering
\includegraphics[width=0.92\textwidth]{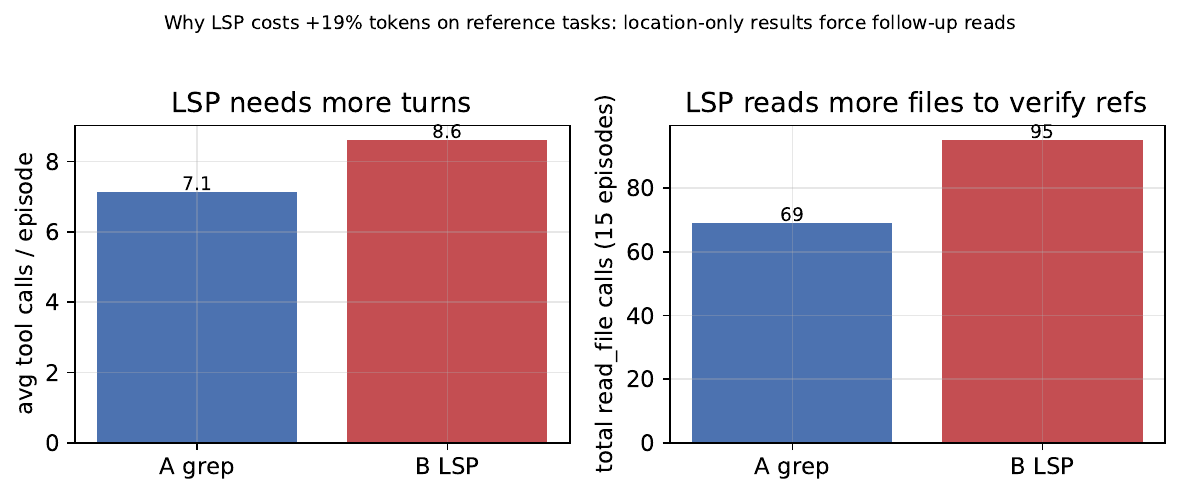}
\caption{\textbf{Why the LSP costs $+19\%$ tokens on reference tasks.} Per-tool-call token cost is nearly
identical between arms ($163$ vs.\ $160$); the difference is \emph{call/turn count}. \emph{Left:} the LSP
arm issues more tool calls per episode. \emph{Right:} it issues far more \code{read\_file} calls.
\code{find\_references} returns \emph{locations only} (\code{path:line}), so the agent must open those
files to verify each call site; \code{grep} returns the matching line's \emph{content inline}, often
needing no follow-up read. Each extra turn re-bills the accumulated conversation context --- so the cost
compounds. An LSP tool that returned the referenced line inline would likely erase the penalty while
keeping precision~$=1.00$ (an actionable H-interface fix).}
\end{figure}

\subsection{Reference-completeness across models, and task-dependent tool choice}

We ran the reference experiment on all three models (60 episodes each). Two results stand out.

\begin{center}\small
\begin{tabular}{@{}lrrcr@{}}
\toprule
Model & A grep F1 & B lsp F1 & $\Delta$F1 & B vs.\ A tokens \\
\midrule
Opus 4.8 & 0.706 & 0.778 & $+0.072$ & $+19\%$ \\
Sonnet 4.6 & 0.706 & 0.789 & $+0.083$ & $+12\%$ \\
Haiku 4.5 & 0.706 & 0.719 & $+0.013$ & $\mathbf{-7\%}$ \\
\bottomrule
\end{tabular}
\end{center}

First, the accuracy benefit is \textbf{model-independent}: the LSP lifts F1 for all three models, with
precision $\to 1.00$ for the two capable models and $0.93$ for Haiku. Second, the \emph{token} picture
is milder than localization: the premium shrinks sharply relative to the localization tax (Sonnet
$+118\%$ on localization $\to +12\%$ here; Haiku $-26\% \to -7\%$) but stays slightly positive for the
two strong models. Unlike localization, here the premium \emph{buys real F1}. A clean token saving
appears only for the weakest model (Haiku, $-7\%$).

\begin{figure}[h]
\centering
\includegraphics[width=0.72\textwidth]{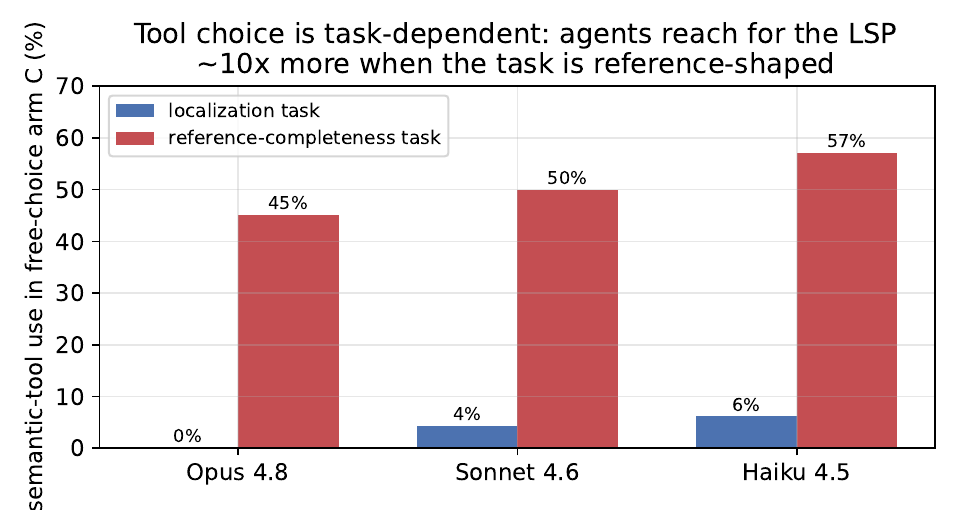}
\caption{\textbf{Tool choice is task-dependent --- the agent has latent routing competence.} Semantic-tool
use in the free-choice arm C, by model, for the two task classes. On \emph{localization} every model
defaults to \code{grep} (0--6\%); on \emph{reference-completeness} the same models reach for the LSP
roughly half the time, \emph{unprompted} (45--57\%). The grep preference is not absolute --- it is
task-shaped. The policy already partially recognizes when ``find all callers'' calls for
\code{find\_references}.}
\end{figure}

This second result is the most important refinement in the study. Our localization data suggested a
\emph{universal} grep preference (arm C used the LSP $0\%$/$4\%$/$6\%$ of the time). The
reference-completeness data overturns the universality: the \emph{same} agents, given the \emph{same}
free choice, use semantic tools $45\%$/$50\%$/$57\%$ of the time when the task is reference-shaped
(Figure~5). So the grep default is \textbf{task-dependent, not a fixed bias} --- the policy already has
substantial latent routing competence and exercises it when the task obviously suits semantic
navigation. This strengthens, rather than weakens, the case for an adaptive router \emph{and} for
training the routing into the policy: the signal is demonstrably already there to be reinforced.

\subsection{What actually determines the LSP's value: lexical noise, not language}

The reference experiment so far compared one Python repo (\code{requests}) and one TypeScript repo
(\code{remeda}), and found the LSP helped on the former ($\Delta$F1 $+0.072$) but not the latter
($+0.000$). That invites the wrong conclusion --- ``TypeScript doesn't benefit from semantic
retrieval.'' It is confounded: \code{remeda} is not just TypeScript, it is a \emph{clean} codebase
(distinctive function names, imported-and-called consistently), so \code{grep} already achieved
precision $1.00$ and the LSP had nothing to add. To separate \emph{language} from \emph{codebase
naming noise}, we added a third repo: \code{hono}, a TypeScript web framework with deliberately
\emph{noisy} target names (\code{html}, \code{stream}, \code{parseAccept}) that collide with comments,
strings, tests, and similarly-named symbols.

\begin{center}\small
\begin{tabular}{@{}llcrrrr@{}}
\toprule
Repo & Language & grep noise & grep F1 & LSP F1 & $\Delta$F1 & grep precision \\
\midrule
\code{requests} & Python & noisy & 0.706 & 0.778 & $+0.072$ & 0.76 \\
\code{remeda} & TypeScript & clean & 0.774 & 0.774 & $+0.000$ & 1.00 \\
\code{hono} & TypeScript & noisy & 0.451 & 0.697 & $\mathbf{+0.245}$ & 0.51 \\
\bottomrule
\end{tabular}
\end{center}

The result \textbf{groups by noise, not language}. Both noisy repos --- Python \emph{and} TypeScript
--- show a clear LSP advantage ($+0.072$, $+0.245$) with low grep precision; the clean repo shows none.
The two TypeScript repos sit at \emph{opposite extremes} ($+0.000$ vs.\ $+0.245$), separated purely by
naming noise. On the noisiest repo (\code{hono}) the LSP not only gives the largest F1 gain but also
\emph{saves} tokens ($-12\%$): when grep is flooded with false positives, the agent reads more to filter
them, so semantic retrieval is strictly better.

The cleanest evidence is \emph{within} a single repo, which eliminates any cross-repo or cross-language
confound entirely. Ranking \code{hono}'s six targets by grep precision:

\begin{center}\small
\begin{tabular}{@{}lcc@{}}
\toprule
Target & grep precision & LSP benefit ($\Delta$F1) \\
\midrule
\code{parseAccept} & 0.07 & $+0.550$ \\
\code{stream} & 0.25 & $+0.428$ \\
\code{html} & 0.18 & $+0.265$ \\
\code{getRuntimeKey} & 0.71 & $+0.150$ \\
\code{escapeToBuffer} & 0.84 & $+0.080$ \\
\code{decodeBase64} & 1.00 & $+0.000$ \\
\bottomrule
\end{tabular}
\end{center}

As grep's precision falls, the LSP's benefit rises, nearly monotonically --- same repo, same language,
same model. \code{decodeBase64} (grep precision $1.00$) gets zero benefit; \code{parseAccept} (grep
precision $0.07$, essentially drowned in false positives) gets $+0.550$. Pooling all targets across all
three repos (Figure~6) gives a single descending relationship (slope ${\approx}-0.49$): every point,
regardless of language or repo, falls on the same line.

\begin{figure}[h]
\centering
\includegraphics[width=0.7\textwidth]{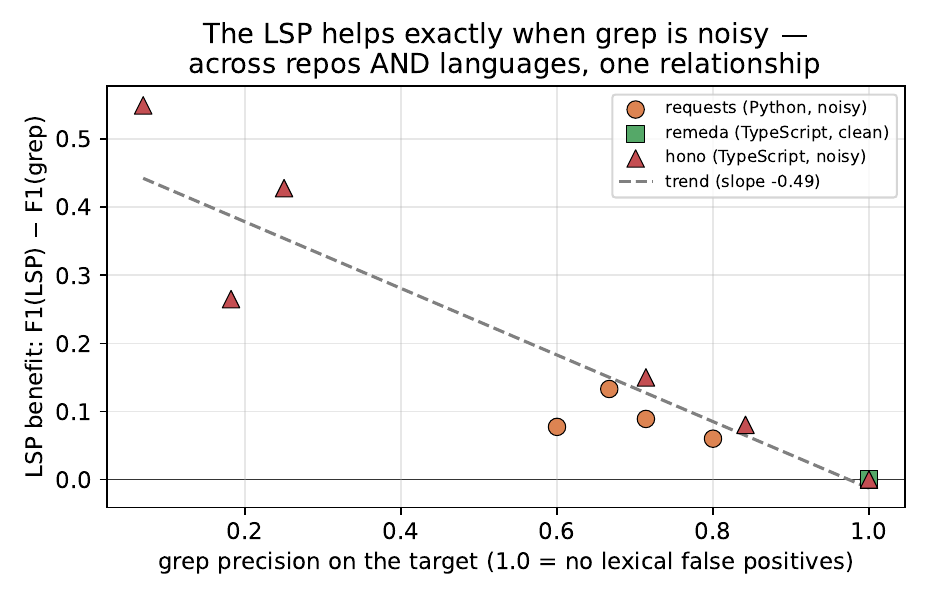}
\caption{\textbf{The LSP helps exactly when \code{grep} is noisy --- one relationship across repos and
languages.} Each point is one target function: x-axis is \code{grep}'s precision on that target (1.0 =
no lexical false positives), y-axis is the LSP's F1 benefit over \code{grep}. Points from all three
repos (Python and TypeScript, clean and noisy) lie on a single descending trend: the LSP's value is
determined by the target's \emph{lexical collision rate}, not by the programming language or its type
system. The clean TypeScript repo (\code{remeda}, green) clusters at precision~$1.0$ / benefit~$0$;
the noisy TypeScript repo (\code{hono}, red) spreads across the high-benefit region exactly where
\code{grep}'s precision is low.}
\end{figure}

This is the study's sharpest causal claim, and it resolves the apparent ``TypeScript doesn't benefit''
result: \textbf{whether semantic retrieval helps a reference-finding task is governed by the symbol's
naming collision rate, not by the language.} The practical rule for an adaptive router follows
directly: route to the LSP when the identifier is lexically ambiguous (a cheap \code{grep} can estimate
its own noise), and stay lexical when it is distinctive.

\subsection{Synthesis}

Across two task classes and three models, a single conditional structure holds:

\begin{quote}
\textbf{``Does an LSP save tokens?'' --- in general, no.} On symbol-named \emph{localization} it
\emph{costs} tokens (a tax that grows with model strength). On \emph{reference-completeness} it does not
save tokens either (a smaller $+12$--$19\%$ premium for capable models) but the premium now buys
\emph{precision} ($1.00$ vs.\ $0.76$) and cannot fix the recall ceiling. It saves tokens only for the
\emph{weakest} model, as a crutch against lexical noise. And the agent's tool choice is
\textbf{task-dependent}: it defaults to \code{grep} on localization ($0$--$6\%$ semantic use) but reaches
for the LSP about half the time on reference tasks ($45$--$57\%$) --- the grep preference is not a fixed
bias but a learned, task-shaped policy.
\end{quote}

This argues against ``LSP-always'' and for \textbf{(a)} an adaptive router keyed on task class, model
capability, and lexical-noise level, and \textbf{(b)} training the \emph{when-to-go-semantic} competence
into the policy --- because handing the model the tool is empirically insufficient.

\subsection{Edit tasks with real test execution}\label{sec:edit}

Localization and reference-completeness are \emph{retrieval} tasks scored against a static oracle. What
matters most for an agent is whether the LSP helps it \emph{edit code correctly}, where correctness is
defined by \emph{execution}. Because the SWE-bench Docker images do not run under emulation on the
available arm64 host, we built a local, SWE-bench-style harness on modern \code{requests} that runs
natively (Python 3.11): the agent edits the working tree via an \code{edit\_file} tool, then the task's
test patch is applied and the target tests run. Resolution is real \code{pass@1}.

\textbf{Single-file edits (6 real bugfix commits).} Grep-only leads (\code{pass@1} $0.83$); adding (C,
$0.58$) or forcing (D, $0.75$) the LSP does not help, and the LSP again carries a token premium.
Failures are \emph{wrong-rewrite} and \emph{no-edit}, never missed-site --- but a single-file task
\emph{cannot} exhibit a cross-file recall miss, so it cannot test whether the LSP's recall ceiling
propagates into edit failures. That needs multi-file edits.

\textbf{Multi-file renames (6 helpers used across 2--6 files).} The agent must update every call site; a
single miss leaves the old name dangling (caught by \code{import requests} or a residual-name grep). We
also measure \emph{site-recall} = fraction of referencing files fully converted. The prediction is
confirmed:

\begin{center}\small
\begin{tabular}{@{}lrrr@{}}
\toprule
Arm & \code{pass@1} & site-recall & missed-site \\
\midrule
A grep-only            & \textbf{1.00} & 1.000 & 0.00 \\
B lsp (location-only)  & 0.67 & 0.930 & 0.33 \\
F lsp (text inline)    & \textbf{0.83} & 0.958 & 0.17 \\
\bottomrule
\end{tabular}
\end{center}

Two confounds had to be removed first, and isolating them is what made the mechanism precise.
\emph{(i) Backend completeness:} \code{pylsp} (jedi) returns only ${\sim}\tfrac13$ of cross-file
references on dynamic Python (5 of 15 for \code{to\_native\_string} vs.\ \code{pyright}'s 14); an LSP
that under-reports references fails a rename regardless of formatting, so we use \code{pyright}.
\emph{(ii) Index warmup:} \code{pyright} resolves references lazily per file, so a cold
\code{find\_references} issued \emph{at a definition} --- the natural query for a rename --- returns only
the definition ($1$ result) until the referencing files are opened; a fixed wait does not help ($1$ after
$40$\,s), but opening all source files does ($1\!\to\!14$). \textbf{An LSP-backed agent that does not warm
the index silently gets incomplete references on exactly the queries a refactor depends on.} This warmup
is not free --- opening every file and letting the server settle costs ${\sim}15$\,s per episode even on
\code{requests} ($19$ files), growing with repo size --- and it is \emph{not} \code{pyright}-specific: a
cold \code{textDocument/references} at the definition of \code{remeda}'s \code{purry} (used across $108$
TypeScript files) likewise returns $1$ until indexing completes (${\sim}10$\,s). Production LSP-MCP servers
treat this as table stakes: Serena ships a build-time \code{serena project index} that persists symbols to
an on-disk cache, \emph{and} a runtime guard \code{\_wait\_for\_cross\_file\_references\_if\_needed()} whose
comment notes that some servers ``require waiting [\ldots] before they can return accurate cross-file
results [\ldots] after at least one file was opened.'' A warmed, ideally pre-cached, index is thus a
prerequisite for any LSP-backed agent, not optional tuning.

With a complete, warmed backend we isolate the practical question: does returning each reference's line
\emph{inline} (as production LSP-MCP servers such as Serena do) beat the bare-location default? Arm F
attaches $\pm 2$ source lines to each \code{pyright} reference. \textbf{It helps substantially and via the
predicted mechanism}: vs.\ B, \code{pass@1} $0.67\!\to\!0.83$, site-recall $0.930\!\to\!0.958$, tokens
$-19\%$, and file-reads/episode $15.2\!\to\!3.2$ (below grep's $4.3$) --- the location-only arm wastes its
turns reading each file back to see the call site. \textbf{Yet even F still loses to grep, and the
residual is fundamental}: B and F both fail the \code{default\_hooks} rename by missing the same site, a
\emph{comment} (\code{\# \ldots default\_hooks() return type}). \code{pyright} returns the six real
references but never the comment --- by design, a comment is not a semantic reference. A rename is a
\emph{textual} operation; semantic references are a strict subset of textual occurrences, so
\code{find\_references} cannot match grep's completeness on edits spanning non-code text. For editing,
grep is the better default; if an LSP is used, a warmed index and inline-context results are table stakes
that recover most --- but not all --- of grep's edge.

\section{Limitations and threats to validity}\label{sec:limits}

The results are \textbf{preliminary}: \textbf{(i)~few repositories, small N} ---
\code{requests} plus \code{remeda}/\code{hono} (TypeScript), small libraries the models likely saw in
pretraining; 6 localization tasks, 5--6 reference targets, and 6 edit / 6 rename tasks, 2--3 rollouts each.
Directions are robust; effect sizes (e.g.\ Sonnet's $+118\%$) are preliminary and will move with more data.
\textbf{(ii)~The edit results are local, not SWE-bench-scored}: SWE-bench Docker images do not run under
emulation on the available arm64 host, so edit tasks use a locally built set (real \code{requests} bugfix
commits and constructed multi-file renames) run natively. This is sound for the \emph{relative} arm
comparison but the \code{pass@1} values are not comparable to a standard SWE-bench leaderboard.
\textbf{(iii)~Oracle quality} drove a mid-study change from
\code{pylsp} to \code{pyright} (jedi was too incomplete to be ground truth) --- itself a finding, and a
caution that any single LSP-as-truth inherits that server's blind spots. \textbf{(iii)~Task coverage}
--- we did not run the \emph{edit} tasks end-to-end with test execution, arm E (static repo-map), or
large repositories where per-symbol LSP cost and the static-index trade-off would bite.
\textbf{(iv)~Language/server} --- Python with \code{pylsp}/\code{pyright}; the originally-motivating
\textbf{TypeScript} case (statically typed, most reliable references) is untested and is the most
important next stratum. \textbf{(v)~Harness specificity} and \textbf{(vi)~token accounting} (we count
total context tokens; caching can change billed cost). \textbf{(vii)~Provenance} as in \S2.

\section{Conclusion}

``Does an LSP save tokens'' is not a tooling opinion --- it is a measurable statement about retrieval
precision under a token budget, and it has gone unmeasured in public while being near-universally
asserted. We reduced it to a single metric (\emph{tokens-to-success at iso-accuracy}), a five-arm
ablation, and a mapping of three failure modes onto movable variables --- and ran a preliminary study that turns the
prior into first numbers. The answer is \textbf{conditional, and in the common case negative}: on
symbol-named localization the LSP \emph{costs} tokens (a tax growing with model strength); on
reference-completeness it buys \emph{precision}, not token savings, at a $\sim$19\% premium and cannot
lift the recall ceiling; it nets a saving only for the weakest model. Underneath sits the most robust
finding: \textbf{every model defaults to \code{grep} when the LSP is merely available.} That is why the
durable solution is not ``add an LSP'' but to make the routing --- and eventually the semantic competence
itself --- \textbf{native to the policy rather than a brittle external layer}, the direction
RL-from-language-server-feedback already points. The contribution is to replace an assertion with a
measurement, and a blanket recommendation with a conditional one. The honest next step is scale: more and
larger repositories, real SWE-bench-scored edits on Docker-capable hardware, and a second model family.

\section*{Code and data availability}
All code (harness, task builders, verifiers, analysis) and raw per-episode results are public at
\url{https://github.com/Poytr1/lsp-vs-grep-token-study}. Key result files: \code{harness/runs/v1\_all\_arms.jsonl}
(localization), \code{ref\_opus.jsonl}/\code{ref\_ts\_*.jsonl} (reference-completeness),
\code{edit\_opus\_verified.jsonl} (single-file edits), and \code{rename\_opus\_verified.jsonl} /
\code{rename\_warm\_BF\_verified.jsonl} (multi-file renames; location-only vs.\ text-inline). Each arm is a
tool surface behind one identical agent loop; figures regenerate via \code{harness/make\_figs.py}.

\section*{How to cite}
\begin{verbatim}
@misc{xu2026lsptokens,
  title  = {Does a Language Server Save Tokens for Coding Agents?
            A Measurement Methodology and Preliminary Study},
  author = {Xu, Pengcheng},
  year   = {2026},
  note   = {Preprint},
  howpublished = {\url{https://github.com/Poytr1/lsp-vs-grep-token-study}}
}
\end{verbatim}

\appendix
\section{Source provenance}
Sources were handled in two tiers. \emph{Fetch-verified (F)} --- retrieved and read this study:
Serena~\cite{serena}, mcp-language-server~\cite{mcpls}, Aider repo-map~\cite{aidermap}, and
arXiv:2604.18413~\cite{tsindex}, 2510.22907~\cite{rlcsf}, 2606.11864~\cite{corebench},
2605.18747~\cite{codeharness} (full abstracts captured). \emph{Established background (K)} --- relied on
from training knowledge and never used for a novel quantitative claim: SWE-bench~\cite{swebench},
the SWE-agent agent-computer-interface~\cite{sweagent}, AutoCodeRover~\cite{autocoderover}, and
Moatless~\cite{moatless}.

\end{document}